\documentclass[lettersize,journal]{IEEEtran}
\usepackage{amsmath,amsfonts}
\usepackage{algorithmic}
\usepackage{algorithm}
\usepackage{array}
\usepackage[caption=false,font=normalsize,labelfont=sf,textfont=sf]{subfig}
\usepackage{textcomp}
\usepackage{stfloats}
\usepackage{url}
\usepackage{verbatim}
\usepackage{graphicx}
\usepackage{cite}
\usepackage{nicematrix}
\usepackage{hhline}
\usepackage{tabularx}
\usepackage{multirow}
\usepackage{changepage}
\usepackage{setspace}
\usepackage{makecell}

\usepackage{amsmath}
\usepackage{booktabs}
\usepackage[hidelinks]{hyperref}
\usepackage[numbers]{natbib}
\usepackage{enumitem}
\usepackage[utf8]{inputenc}

\begin{document}

\title{TrajFusionNet: Pedestrian Crossing Intention Prediction via Fusion of Sequential and Visual Trajectory Representations}

\author{François G. Landry and Moulay A. Akhloufi\\
Perception, Robotics and Intelligent Machines Research Group (PRIME)\\
Department of Computer Science, Université de Moncton, Canada\\
{\tt \{efl7126, moulay.akhloufi\}@umoncton.ca}
}



\maketitle

\begin{abstract}
With the introduction of vehicles with autonomous capabilities on public roads, predicting pedestrian crossing intention has emerged as an active area of research. The task of predicting pedestrian crossing intention involves determining whether pedestrians in the scene are likely to cross the road or not. In this work, we propose TrajFusionNet, a novel transformer-based model that combines future pedestrian trajectory and vehicle speed predictions as priors for predicting crossing intention. TrajFusionNet comprises two branches: a Sequence Attention Module (SAM) and a Visual Attention Module (VAM). The SAM branch learns from a sequential representation of the observed and predicted pedestrian trajectory and vehicle speed. Complementarily, the VAM branch enables learning from a visual representation of the predicted pedestrian trajectory by overlaying predicted pedestrian bounding boxes onto scene images. By utilizing a small number of lightweight modalities, TrajFusionNet achieves the lowest total inference time (including model runtime and data preprocessing) among current state-of-the-art approaches. In terms of performance, it achieves state-of-the-art results across the three most commonly used datasets for pedestrian crossing intention prediction. 
\end{abstract}


\section{Introduction}

Substantial efforts have been devoted to developing methods for pedestrian detection and identifying pedestrians actively crossing \citep{dollar2011pedestrian, gandhi2007pedestrian}. The latter is typically framed as an action recognition problem, where the goal is to infer a class label (e.g., whether a pedestrian is \textit{currently} crossing) based on observed actions. Action prediction, on the other hand, involves anticipating whether an action will occur in the future, making it inherently more challenging. Predicting pedestrian crossing intention is particularly challenging due to the fact that pedestrians are highly dynamic, can change direction quickly \citep{galvao2023pedestrian} and that many factors can change the trajectories of pedestrians \citep{liu_spatiotemporal_2020}. From the ego-vehicle perspective, pedestrians can easily get occluded by other pedestrians or by objects such as other vehicles and structures along the road. Pedestrian movement is highly dependent on other traffic objects in the scene such as other agents (e.g., vehicles, other pedestrians) and traffic elements (e.g., sidewalks, zebra crossings, traffic lights). Despite the difficulty surrounding crossing intention prediction, the task has significant practical applications. Predicting whether a pedestrian is likely to cross the road can enable autonomous vehicles to compute and potentially execute an avoidance plan (e.g., braking or changing direction) in advance, reducing the risk of collisions.

In this work, we introduce TrajFusionNet, a novel model for predicting pedestrian crossing intention. The model leverages pedestrian trajectory and vehicle speed predictions as priors for predicting crossing intention. TrajFusionNet is composed of two branches: a Sequence Attention Module (SAM) and a Visual Attention Module (VAM). In the SAM branch, an encoder-decoder transformer is used to predict the future pedestrian trajectory and vehicle speed. This sequential representation, along with the past observed trajectory, is then used as input to a transformer encoder. Complementarily, the VAM branch enables learning from a visual representation of the predicted pedestrian trajectory by overlaying predicted pedestrian bounding boxes onto scene images. The VAM branch is composed of two VAN (Visual Attention Network \cite{guo2023visual}) blocks. TrajFusionNet achieves state-of-the-art performance across the three most widely used datasets for pedestrian crossing intention prediction. Additionally, by employing a small number of lightweight modalities, TrajFusionNet minimizes total inference time (including model runtime and data preprocessing), outperforming current state-of-the-art approaches.

\section{Related Work}

In this section, we provide a short review of the literature on pedestrian crossing intention prediction. For more extensive surveys, the reader is referred to the following reviews: \cite{LANDRY2024129105, galvao2023pedestrian, zhang2023pedestrian}. The first approaches proposed in the literature for pedestrian crossing intention were based on probabilistic models \cite{bandyopadhyay2013intention, kooij_context-based_2014} and on applying machine learning classifiers to hand-crafted features \cite{kohler2013autonomous, schneemann2016context}. The early deep learning methods involved applying CNNs (convolutional neural networks) to parts of the scene image such as the area surrounding the pedestrian \cite{rasouli_1_are_2017, varytimidis_2_action_2018}. To integrate the time dimension across video frames, authors have proposed using 3D CNN-based models \cite{saleh_3_real-time_2019} or applying RNNs (recurrent neural networks) to sequential features \cite{li_3_pedestrian_2020, lorenzo_8_rnn-based_2020}. 

A variety of features were proposed for use in sequential models, including pedestrian pose keypoints, bounding box coordinates, and vehicle speed. Researchers were able to improve predictive results by leveraging more modalities as input into RNN-based models, usually at the cost of increased inference time \cite{kotseruba_2_do_2020, ranga_6_vrunet_2020}. State-of-the-art results were obtained using hierarchical RNN architectures, which are composed of multiple RNNs where the output of a lower-level RNN is used as the input of a higher-level RNN. One such model is SF-GRU \cite{rasouli_8_pedestrian_2020}, where the different modalities are fused into the network gradually according to their complexity, from the most complex at the bottom to the least complex at the top. Results were further improved by using attention layers to assign more or less importance to frames at certain timestamps (temporal attention) or to areas of an image (visual attention) \cite{kotseruba_4_benchmark_2021, yang_2_predicting_2022}. PCPA \cite{kotseruba_4_benchmark_2021} uses a multi-modal RNN with temporal attention layers as well as modality attention between the different branches of the network. TrouSPI-Net \cite{gesnouin2021trouspi} is a similar model, but with the addition of a branch where 2D body pose keypoints are transformed into a pseudo-image, which is then processed with atrous convolutions followed by channel and spatial attention layers.

The advent of transformers \cite{vaswani_attention_2017} led to an increase in predictive performance compared to other sequential models such as RNNs. Achaji et al. \cite{achaji_3_is_2022} obtained state-of-the-art results by using an encoder-decoder transformer model and using solely bounding box coordinates as input features. Zhou et al. \cite{zhou_pit_2023} propose PIT (Progressive Interaction Transformer), where a transformer layer is applied to each video frame in a temporal sequence, allowing early interactions between modalities. Bai et al. \cite{bai2022deep} propose to use knowledge distillation with a student network composed of a basic transformer encoder using only bounding box coordinates and a teacher network leveraging 3D convolutions and attention layers applied to multiple modalities. Vision transformers, which apply the self-attention mechanism to image patches, have been used for pedestrian intention prediction as part of approaches proposed in \citep{lorenzo_4_capformer_2021, zhao_action-vit_2021}.

Another family of approaches proposed in the literature consists of generative approaches, which tackle the pedestrian intention problem by predicting future representations, which are typically future frames. Gujjar and Vaughan \cite{gujjar_classifying_2019} and Chaabane et al. \cite{chaabane_2_looking_2020} propose to use an encoder-decoder network to predict future frames, using a 3D CNN as the encoder and a convolutional LSTM at the decoder level. An important weakness of generative approaches is their high inference time which is reported as over 100 ms on commodity GPUs. TrajFusionNet draws inspiration from previously proposed generative approaches; however, instead of predicting future image frames as priors, we predict lightweight modalities (future bounding box locations and vehicle speed), allowing to limit inference time. 

Graph-based approaches have been very successful recently in the pedestrian crossing intention literature. Graph-based approaches aim to model spatiotemporal dependencies between elements in the scene with graphs. The most common deep learning model that allows learning from graph structures is the graph convolutional network (GCN). A GCN applied to a scene graph is used by Song et al. \cite{song2022pedestrian}, where each node represents an object in the scene (pedestrian or traffic light) and each edge represents the strength of the interaction between the two nodes. GCN models have also been applied to pedestrian pose keypoints in \cite{cadena2022pedestrian, yang2023dpcian}. In PedAST-GCN, Ling et al. \cite{ling2024pedast} use graph representations that allow to integrate the bounding box and vehicle speed modalities in addition to pose keypoints into GCN layers, which are then followed by attention layers.

Some authors have proposed approaches that leverage multi-task learning, where the model is trained to solve multiple tasks at the same time. As such, pedestrian crossing prediction has been combined  with other tasks such as trajectory prediction \cite{schorkhuber2022feature, ranga_6_vrunet_2020, cao_3_using_2020, sui_joint_2021, achaji_3_is_2022}, target location prediction \cite{schorkhuber2022feature}, time to cross the street \cite{pop_5_multi-task_2019} and predicting additional pedestrian actions (e.g., walking, standing) \cite{yao2021coupling, ranga_6_vrunet_2020, zhao_action-vit_2021, zhai2022social}. Multi-task learning allows the sharing of commonalities across tasks through a shared representation, leading to an efficient use of model parameters. Multi-task models typically contain a backbone of common layers followed by task-specific heads at the end of the network. In other models, the output from one subtask is used as the input of another task \citep{achaji_3_is_2022, cao_3_using_2020}. TrajFusionNet follows this approach by leveraging pedestrian trajectory and vehicle speed predictions as priors for predicting crossing intention.

\section{Method}

\subsection{Problem Formulation}

The problem at hand consists of predicting whether a pedestrian is going to cross (or not cross) during a video sequence, given a set of observation frames, and for all pedestrians detected. The pedestrian crossing intention problem can be formulated as follows. It consists of a binary classification task at time \begin{math}t\end{math} where crossing action \begin{math}A_i \in \{ 0, 1 \}\end{math} is predicted in the future for pedestrian \begin{math}i\end{math} given previous observations \begin{math}\boldsymbol{M}_i\end{math}:

\begin{equation*}
\mathrm{\boldsymbol{M}}_i = \{ \mathrm{\boldsymbol{m}}^{t-m}_i, \mathrm{\boldsymbol{m}}^{t-m+1}_i,...,\mathrm{\boldsymbol{m}}^{t}_i \}
\end{equation*}

We follow the benchmark parameters proposed by Kotseruba et al. \cite{kotseruba_4_benchmark_2021} and predict whether the pedestrian will cross between \begin{math}1\end{math} and \begin{math}2\end{math} seconds after time \begin{math}t\end{math}. 

\subsection{Input Modalities}

Our model uses three input modalities: a sequence of pedestrian bounding boxes, a sequence of vehicle speeds, and a sequence of scene images. As such, \begin{math}\boldsymbol{M}_i\end{math} can be divided into three tensors:
\begin{math}
\mathrm{\boldsymbol{B}}_i = \{ \mathrm{\boldsymbol{b}}^{t-m}_i, \mathrm{\boldsymbol{b}}^{t-m+1}_i,...,\mathrm{\boldsymbol{b}}^{t}_i \}
\end{math}, the bounding box sequence,
\begin{math}
\boldsymbol{V} = \{ v^{t-m}, v^{t-m+1}, ..., v^{t} \}
\end{math}, the vehicle speed sequence, and \begin{math}
\mathrm{\boldsymbol{\mathcal{I}}} = \{ \mathrm{\boldsymbol{I}}^{t-m}, \mathrm{\boldsymbol{I}}^{t-m+1},...,\mathrm{\boldsymbol{I}}^{t} \}
\end{math}, the scene video sequence. We keep 4 values for each element in the bounding box sequence, namely the \begin{math}x\end{math} and \begin{math}y\end{math} coordinates for the upper-left and bottom-right corners. The vehicle speed is computed differently depending on the dataset used for evaluating the model. The PIE dataset \citep{rasouli_pie_2019} provides raw vehicle speed values, which are directly used as input to the model. For the JAAD dataset \citep{rasouli_1_are_2017}, only vehicle speed categories are provided. We encode the categories ordinally as follows: \begin{math}\{0\end{math}: stopped, \begin{math}1\end{math}: decelerating, \begin{math}2\end{math}: moving slow, \begin{math}3\end{math}: moving fast, \begin{math}4\end{math}: accelerating\begin{math}\}\end{math}.

\subsection{Architecture}

\begin{figure*}
    \begin{center}
        \includegraphics[width=0.8\linewidth]{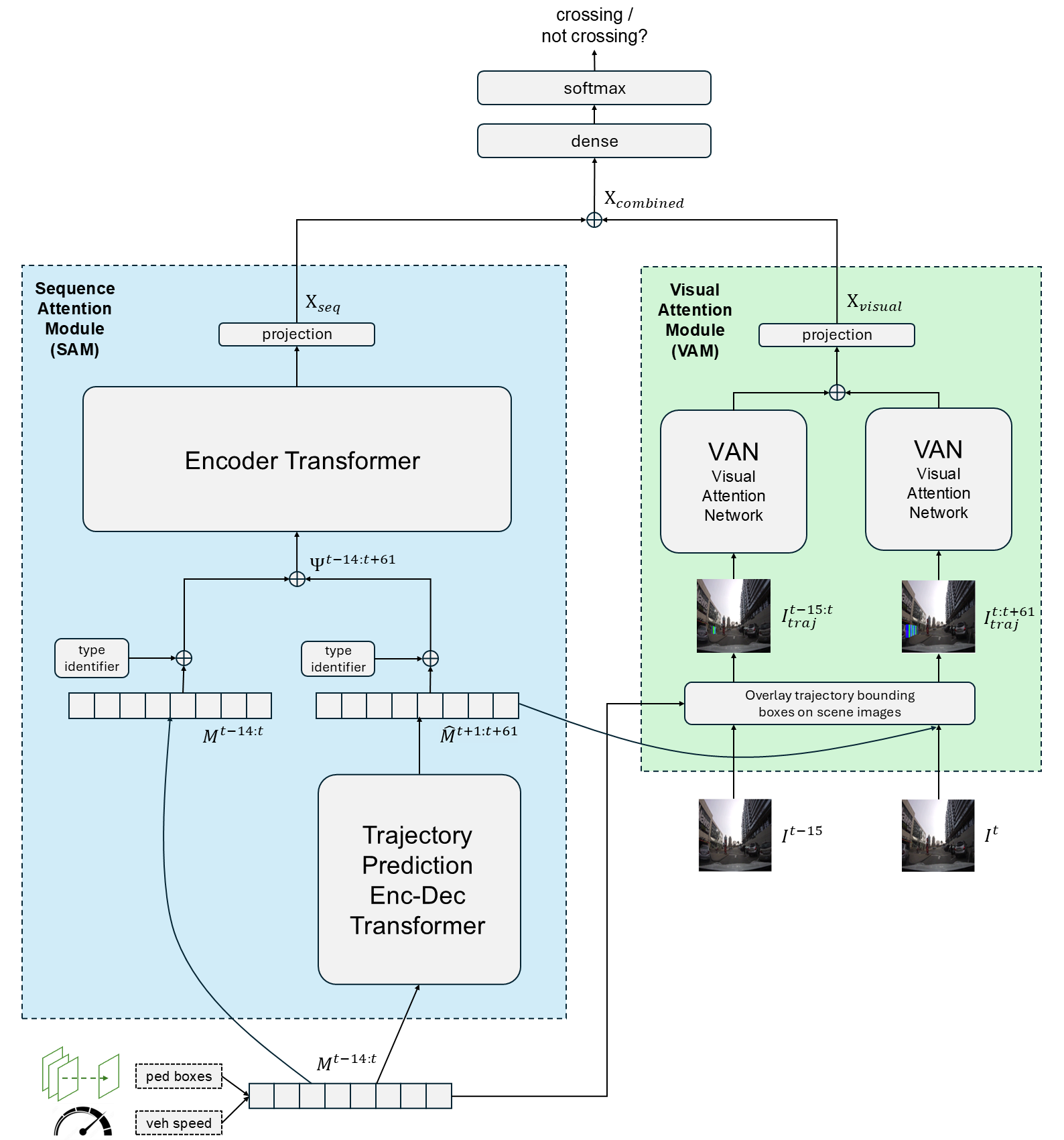}
    \end{center}
   \vspace*{-10pt}
   \caption{TrajFusionNet architecture. The model is composed of two branches: a Sequence Attention Module (SAM) and a Visual Attention Module (VAM). In the SAM branch, an encoder-decoder transformer is used to predict the future pedestrian trajectory and vehicle speed. This sequential representation, along with the past observed trajectory and vehicle speed, is then passed as input to a transformer encoder. In the VAM branch, the input consists of observed and predicted pedestrian bounding boxes overlaid onto scene images. Two instances of the Visual Attention Network (VAN) \cite{guo2023visual} are used: one instance is applied to the first video frame in the observation period, overlaid with the observed pedestrian bounding boxes, and the second instance is applied to the last frame, overlaid with the predicted pedestrian bounding boxes. The outputs from the VAM and SAM branches are passed through projection layers and then combined with dense layers in a late-fusion fashion.}
\label{fig:figure1_architecture}
\end{figure*}

The proposed architecture is shown in Figure~\ref{fig:figure1_architecture}. The model is composed of two branches: a \textit{Sequence Attention Module (SAM)} and a \textit{Visual Attention Module (VAM)}. The two branches are merged into a late-fusion fashion with a dense layer.

\textbf{Sequence Attention Module (SAM)}: The role of the SAM branch is to extract insights by applying the attention mechanism to a sequential representation of past observed and predicted pedestrian locations and vehicle speed. The SAM branch is composed of two transformer blocks. The first transformer is an encoder-decoder transformer that is used to perform trajectory prediction to estimate future pedestrian bounding boxes and future vehicle speed. The trajectory is predicted over the next 60 frames (up until 2 seconds after time \begin{math}t\end{math}). The trajectory prediction tensor, \begin{math}\boldsymbol{\hat{M}}^{t+1:t+61}\end{math}, is then concatenated with the past trajectory tensor, \begin{math}\boldsymbol{M}^{t-14:t}\end{math}, to form a new tensor, \begin{math}\boldsymbol{\psi}^{t-14:t+61}\end{math}. \begin{math}\boldsymbol{\psi}^{t-14:t+61}\end{math} is fed to the input of an encoder-only transformer, whose task is to output an encoding to be used for classification.

\textbf{Trajectory Prediction Transformer}: The trajectory prediction transformer consists of a non-autoregressive encoder-decoder transformer. The future trajectories are all predicted in one pass (i.e., the previously generated token is not required to generate the next token), which allows to reduce the inference time considerably. When predicting trajectory, we predict both the future bounding box coordinates and the future vehicle speed.

We use the non-autoregressive encoder-decoder transformer implementation provided in the TSLib time series library \citep{wang2024deep}. The input to the encoder consists of the past observed trajectory sequence, \begin{math}\boldsymbol{M}^{t-14:t} \in \mathbb{R}^{d_{\text{seq}} \times m} \end{math}, where \begin{math}d_{seq}\end{math} is the length of the past trajectory (15 frames) and \begin{math}m\end{math} is the input modalities dimension (\begin{math}m=5\end{math}). For the pedestrian bounding box modality, all coordinates in the sequence are offset by subtracting the coordinates of the first bounding box at time \begin{math}t-15\end{math}. The input to the decoder consists of the concatenation of the past trajectory sequence, \begin{math}\boldsymbol{M}^{t-14:t} \in \mathbb{R}^{d_{\text{seq}} \times m} \end{math}, with an empty tensor, \begin{math}\mathbf{0} \in \mathbb{R}^{d_{\text{pred}} \times m} \end{math}, where \begin{math}d_{pred}\end{math} is the length of the predicted trajectory (60 frames). At the end of the decoder layer stack, a projection layer transforms the output into a tensor of dimensions \begin{math}(d_{\text{seq}} + d_{\text{pred}}) \times m\end{math}, where \begin{math}d_{seq}+d_{pred}\end{math} represents the total length of the past and predicted sequences (75 frames). At the output of the decoder, we only keep the predicted sequence (last 60 frames), leading to the trajectory prediction tensor \begin{math}\boldsymbol{\hat{M}}^{t+1:t+61}\in \mathbb{R}^{d_{\text{pred}} \times m}\end{math}. 

\textbf{Sequence Type Identifiers}: Before sending the past observed trajectory tensor, \begin{math}\boldsymbol{M}^{t-14:t}\end{math}, and the predicted trajectory tensor, \begin{math}\boldsymbol{\hat{M}}^{t+1:t+61}\end{math}, to the subsequent encoder transformer, we augment the two tensors with sequence type identifiers. Sequence type identifiers encode whether sequence tokens consist of past trajectory values or rather of trajectory values predicted in the future. This helps attention heads to attend specifically to past trajectory tokens and to ignore future trajectory tokens, or vice versa. We simply append a fixed scalar of \begin{math}0\end{math} to past trajectory tokens and a fixed scalar of \begin{math}1\end{math} to future trajectory tokens, such that:

\begin{itemize}[leftmargin=.3in]
  \item \begin{math}\boldsymbol{M}^{t}\end{math} is augmented as \begin{math}[\boldsymbol{M}^{t}, 0]\end{math}
  \item \begin{math}\boldsymbol{\hat{M}}^{t}\end{math} is augmented as \begin{math}[\boldsymbol{\hat{M}}^{t}, 1]\end{math}
\end{itemize}

The resulting tensors are concatenated to form a new tensor, \begin{math}\boldsymbol{\psi}^{t-14:t+61}\end{math}. \begin{math}\boldsymbol{\psi}^{t-14:t+61}\end{math} is then fed to the input of an encoder-only transformer whose task is to output an encoding to be used for classification. A final projection layer is added at the end of the SAM branch.

\textbf{Visual Attention Module (VAM)}: The role of the VAM branch is to extract insights by applying the attention mechanism to a visual representation of the pedestrian trajectories as well as the surrounding contextual scene. The VAM branch is composed of Visual Attention Networks (VANs) \cite{guo2023visual}. The VAN network is based on large kernel attention (LKA), which is composed of three components: a spatial local convolution, a spatial long-range convolution, and a channel convolution. The LKA combines the advantages of convolution and self-attention, and allows to better capture long-range relationships. In the context of pedestrian crossing intention, long-range relationships may emerge from interactions between pedestrians or vehicles that are positioned far apart in the ego view.

The VAN network is applied to full scene images captured from the ego-vehicle. These images are augmented with colored rectangles that correspond to the pedestrian bounding boxes associated with the past observed and predicted trajectories. This allows the VAN network to capture interactions between the pedestrian trajectory and the surrounding context. Two instances of VAN are used:

\begin{itemize}
  \item The first VAN instance is applied to the first video frame available in the observation period, \begin{math}\boldsymbol{I}^{t-15}\end{math}. Rectangles corresponding to the observed pedestrian bounding boxes (\begin{math}\boldsymbol{M}^{t-14:t}_{1:4}\end{math}) are added to the image, forming \begin{math}\boldsymbol{I}^{t-15:t}_{traj}\end{math}, which is used as input into the first VAN network.
  \item The second VAN instance is applied to the last frame available in the observation period, \begin{math}\boldsymbol{I}^{t}\end{math}. Rectangles corresponding to the predicted pedestrian bounding boxes (\begin{math}\boldsymbol{\hat{M}}^{t+1:t+61}_{1:4}\end{math}) are added to the image, forming \begin{math}\boldsymbol{I}^{t:t+61}_{traj}\end{math}, which is used as input into the second VAN network.
\end{itemize}

\begin{figure}
    \begin{center}
        \includegraphics[width=1\linewidth]{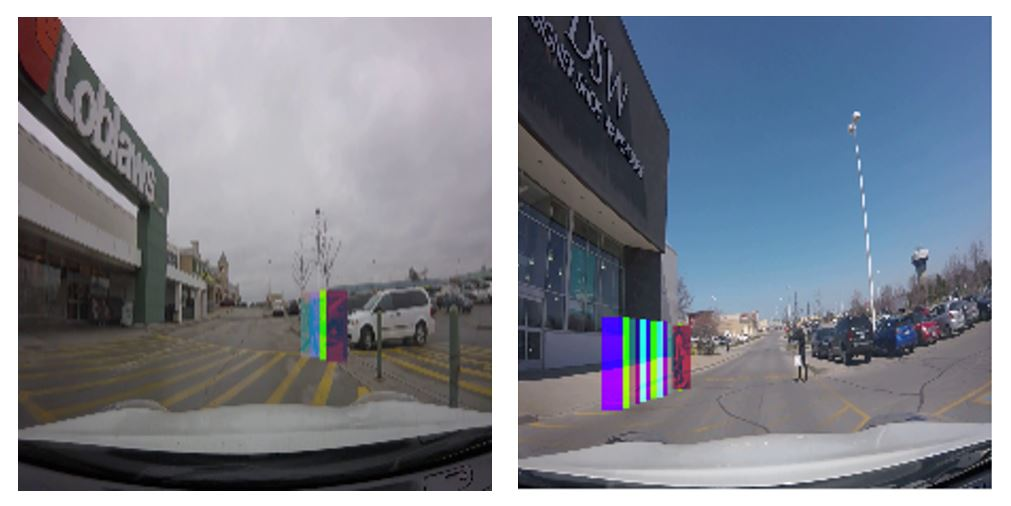}
    \end{center}
   \vspace*{-10pt}
   \caption{Scene images augmented with trajectory boxes. The image on the left corresponds to the first frame available in the observation period augmented with observed pedestrian bounding boxes (\begin{math}\boldsymbol{I}^{t-15:t}_{traj}\end{math}). The image on the right corresponds to the last frame available in the observation period augmented with predicted pedestrian bounding boxes (\begin{math}\boldsymbol{I}^{t:t+61}_{traj}\end{math}).}
\label{fig:figure2_trajectory_van}
\end{figure}

Rectangles corresponding to the pedestrian trajectory bounding boxes are added to two channels only (blue and green channels in the BGR image). One channel is kept so that the pedestrian appearance can still be leveraged by the network. The colors used for drawing the rectangles come from the ADE20k dataset color palette used for scene segmentation \cite{zhou2017scene}. Examples of scene images augmented with trajectory bounding boxes are shown in Figure~\ref{fig:figure2_trajectory_van}. The output from the two VAN networks are concatenated and then fed into a projection layer whose output is to be used for classification.

\textbf{Late Fusion}: At the end of the network, the outputs from the Sequence Attention Module (SAM) and the Visual Attention Module (VAM) are merged in a late-fusion fashion using dense layers of sizes 80, 40, and 2.

\section{Evaluation}

\subsection{Datasets}

Two datasets are used to evaluate our model on predicting pedestrian crossing intention: JAAD (Joint Attention for Autonomous Driving) \cite{rasouli_1_are_2017} and PIE (Pedestrian Intention Estimation) \cite{rasouli_pie_2019}. These two datasets are the most popular datasets used in the literature for the prediction of pedestrian crossing intention.

\textbf{JAAD}: The JAAD dataset \citep{rasouli_1_are_2017} is a naturalistic dataset providing 346 clips of pedestrians prior to crossing events. It was filmed in North America and Europe and under varying weather conditions. The dataset provides ground truth bounding boxes for pedestrians as well as behavioral tags describing pedestrian actions. There are also behavioral tags assigned to the vehicle driver, although a numerical value for vehicle speed is not provided. The JAAD dataset has been divided into two subsets \citep{kotseruba_4_benchmark_2021}: JAAD{\scriptsize beh}, which is skewed towards pedestrians who are crossing or are about to cross, and JAAD{\scriptsize all}, which consists of the complete dataset and contains an additional 2100 visible pedestrians who are away from the road and are not crossing.

\textbf{PIE}: The same authors who proposed the JAAD dataset compiled a second dataset called PIE (Pedestrian Intention Estimation) \citep{rasouli_pie_2019}. PIE is a larger dataset (with almost 10 times the number of frames) and contains longer pedestrian clips and better ego-vehicle information (including speed, GPS location, and heading angle). Compared to JAAD, pedestrians are more diverse in appearance, in the type of behavior they exhibit, and in their location with respect to the curb. However, the PIE dataset was filmed only in Toronto, Canada in clear weather, making it less diverse in terms of driving context.

\subsection{Training and Model Settings}

\textbf{Modular Training}: Given the modularity and relative size of our model, we train it in steps, where each module is pretrained separately. After pretraining the lower modules, we freeze their weights and proceed to fine-tune the subsequent modules. To train each module, we append a classification head composed of two dense layers (with 40 and 2 neurons, respectively), which then gets removed when the trained module is added to the rest of the model.

This approach resembles greedy layer-wise training \cite{hinton2006fast, bengio2006greedy}, with the modification that it is not individual layers but groups of layers (within each module) that are trained together before moving on to the next group. Layer-wise training addresses two challenges: it mitigates overfitting by enhancing the generalization of learned representations in the early layers, and it helps prevent gradient vanishing during backpropagation \cite{barshan2015stage}.

\textbf{Training Procedure and Learning Objectives}:
The first module that we pretrain is the trajectory prediction encoder-decoder transformer. This transformer is trained to predict the pedestrian bounding box and vehicle speed over the next 60 frames (\begin{math}\boldsymbol{\hat{M}}^{t+1:t+61}\end{math}). To train the trajectory prediction transformer, we extract overlapping sequences from pedestrian tracks, with the overlap factor being equal to the one suggested in the benchmark proposed by Kotseruba et al. \cite{kotseruba_4_benchmark_2021}. For the PIE dataset, this results in 49,527 sequences in the training set. We use a mean square error loss function as the training objective:

\begin{equation*}
L = \frac{1}{NTC}\sum_{i=1}^{N}\sum_{j=1}^{T}\sum_{k=1}^{C}{\left(y_{ijk}-{\hat{y}_{ijk}} \right)}^2
\end{equation*}
where \begin{math}N\end{math} is the number of training samples,  \begin{math}T\end{math} is the length of the predicted trajectory (60 frames), and \begin{math}C\end{math} is the number of values predicted as part of the trajectory. The latter includes five z-score normalized values: the \begin{math}x\end{math} and \begin{math}y\end{math} coordinates of the upper-left and bottom-right corners of the bounding box, as well as the vehicle speed.  

Once the trajectory prediction transformer has been trained, we now train the encoder transformer. We freeze the weights of the trajectory prediction transformer and only train the new weights added with the encoder transformer, as well as additional dense layers for classification.  For training, we use the same dataset splits and evaluation parameters (track split, observation length and prediction horizon) as specified in the benchmark proposed by Kotseruba et al. \cite{kotseruba_4_benchmark_2021}. For the PIE dataset, this results in 4,770 sequences in the training set. We use a weighted cross-entropy loss function as the training objective: 

\begin{equation*}
\begin{split}
L = -\frac{1}{N} \sum_{i=1}^{N} \Big( & \, \alpha y_i \log(\hat{y_i}) \\
& + (1 - \alpha)(1 - y_i) \log(1 - \hat{y_i}) \Big)
\end{split}
\end{equation*}

\noindent where \begin{math}N\end{math} is the number of training samples and \begin{math}\alpha\end{math} is a weighting factor to account for class imbalance.

The VAN networks are trained in a similar manner, with additional dense layers for classification. Each VAN is trained independently. The trajectory prediction transformer's weights are frozen, and its output is used to overlay trajectory bounding boxes on the scene images, which serve as input to the VANs.

In the final training step, the complete model is trained with the weights of the SAM and VAM branches frozen. Only the weights of the projection layers at the ends of the branches and the final dense layers are updated.

\textbf{Implementation and Training Parameters}: When conducting experiments, we keep the implementation and training parameters consistent across datasets during training/inference. In the trajectory prediction encoder-decoder transformer, we set the number of encoder layers to 8, the number of decoder layers to 8, the number of attention heads to 4, \begin{math}d_{model}\end{math} to 128, and the dimension of the fully-connected layers to 512. In the subsequent encoder transformer, we set the number of encoder layers to 6, the number of attention heads to 12, \begin{math}d_{model}\end{math} to 128, and the dimension of the fully-connected layers to 1024. When it comes to the two VAN models, we keep the parameters suggested by the authors in the VAN-B2 version \cite{guo2023visual}, which contains 26.6M parameters. The training parameters used for each module of the model are shown in Table~\ref{tab:training_params}. All modules are trained using a linear learning rate decay schedule with an initial warmup phase. The values reported in Table~\ref{tab:training_params} correspond to the peak learning rate attained during training. During the final modular training step (remaining layers), we disable learning in the projection layer of the VAM branch for the first 15 epochs (by setting the learning rate to 0) to improve learning in the SAM branch.

We also train a lightweight version of the TrajFusionNet architecture, which we name TrajFusionNet-Small, and that contains significantly less parameters (5.20M parameters vs. 58.28M parameters for the full TrajFusionNet model). In TrajFusionNet-Small, the dimension of fully-connected layers in both transformers is reduced to 256; the encoder-decoder transformer is downsized to 2 encoder layers, 2 decoder layers, and 2 attention heads; and the encoder transformer is downsized to 2 encoder layers and 2 attention heads. A single VAN instance is applied to the last frame available in the observation period, overlaid with both the observed and predicted bounding boxes. The VAN-B0 version \cite{guo2023visual} of VAN is used, which contains 4.1M parameters.

\textbf{Evaluation Procedure}: We follow the same evaluation procedure as proposed in the benchmark by Kotseruba et al. \cite{kotseruba_4_benchmark_2021}. We use 0.53 seconds for the observation length (16 frames) and predict pedestrian crossing 1 to 2 seconds (30 to 60 frames) after the observation period. We train and test our model on three datasets: PIE, JAAD{\scriptsize all}, and JAAD{\scriptsize beh}. The results reported for each dataset correspond to the model trained and tested on that specific dataset. We provide results using the following common classification metrics: accuracy (Acc), area under the curve (AUC), F1-score (F1), precision (P), and recall (R).

\begin{table}[t]
  \centering
  \caption{Parameters used for training each module of the model during modular training}
  \resizebox{\columnwidth}{!}{%
  \begin{tabular}{l c ccc}
    \toprule
    Module & \makecell{Maximum \\ Learning Rate} & Epochs & Batch Size \\
    \toprule
    \makecell[l]{Trajectory prediction \\ enc-dec TF} & 5e-5 & 40 & 64 \\ \hline 
    Encoder TF & 5e-6 & 60 & 16 \\ \hline
    VAN & 5e-5 & 15 & 16 \\ \hline
    Remaining layers & 5e-6 & 60 & 16 \\
    \bottomrule    
  \end{tabular}
  }
  \label{tab:training_params}
\end{table}

\subsection{Results}

Table~\ref{tab:comparison_sota} shows a comparison of state-of-the-art (SOTA) methods for the prediction of crossing intention. Values shown in bold correspond to the best value obtained across all models, while values that are underlined correspond to the second best value. TrajFusionNet achieves state-of-the-art results, obtaining the highest number of top metrics. It performs particularly well in terms of accuracy and F1 score, achieving the highest accuracy and either the best or second-best F1 score across all datasets. TrajFusionNet also achieves consistently strong results across datasets, indicating that the model is less prone to overfitting on specific datasets.

\begin{table*}[t]
  \centering
  \caption{Comparison of state-of-the-art methods for the prediction of crossing intention on the PIE and JAAD datasets. Values in bold indicate the best performance, while underlined values represent the second-best performance.}
  \resizebox{\textwidth}{!}{%
  \begin{tabular}{lcccccccccccccccc}
    \toprule
    \multirow{2}{*}{Model} & \multirow{2}{*}{Year} & \multicolumn{5}{c}{PIE} & \multicolumn{5}{c}{JAAD{\scriptsize all}} & \multicolumn{5}{c}{JAAD{\scriptsize beh}} \\
    \cmidrule(lr){3-7} \cmidrule(lr){8-12} \cmidrule(lr){13-17}
                                & & Acc & AUC & F1 & P & R & Acc & AUC & F1 & P & R & Acc & AUC & F1 & P & R \\ 
    \toprule
    SF-GRU \cite{rasouli_8_pedestrian_2020} & 2020 & 0.82 & 0.79 & 0.69 & 0.67 & 0.70 & 0.84 & 0.84 & 0.65 & 0.54 & \underline{0.84} & 0.51 & 0.45 & 0.63 & 0.61 & 0.64 \\
    PCPA \cite{kotseruba_4_benchmark_2021} & 2020 & 0.86 & \underline{0.91} & 0.78 & 0.69 & \underline{0.89} & 0.83 & 0.77 & 0.57 & 0.50 & 0.66 & 0.58 & 0.47 & 0.73 & 0.61 & 0.92 \\
    TrouSPI-Net \cite{gesnouin2021trouspi} & 2021 & 0.88 & 0.88 & 0.80 & 0.73 & \underline{0.89} & 0.85 & 0.73 & 0.56 & 0.57 & 0.55 & 0.64 & 0.56 & 0.76 & 0.66 & 0.91 \\
    Yang et al. \cite{yang_2_predicting_2022} & 2022 & 0.89 & 0.86 & 0.80 & 0.79 & 0.81 & 0.83 & 0.82 & 0.63 & 0.51 & 0.81 & 0.62 & 0.51 & 0.76 & 0.63 & \textbf{\textcolor{black}{0.95}} \\
    Pedestrian Graph + \cite{cadena2022pedestrian} & 2022 & 0.89 & 0.90 & 0.81 & 0.83 & 0.79 & 0.86 & \textbf{\textcolor{black}{0.88}} & 0.65 & 0.58 & 0.75 & 0.70 & \textbf{\textcolor{black}{0.70}} & 0.76 & \textbf{\textcolor{black}{0.77}} & 0.75 \\
    Song et al. \cite{song2022pedestrian} & 2022 & \textbf{\textcolor{black}{0.92}} & \underline{0.91} & \textbf{\textcolor{black}{0.86}} & 0.82 & \textbf{\textcolor{black}{0.90}} & 0.87 & 0.81 & 0.65 & 0.60 & 0.71 & 0.68 & 0.60 & 0.78 & 0.68 & 0.91 \\
    Bai et al. \cite{bai2022deep} & 2022 & 0.89 & 0.88 & 0.79 & 0.74 & 0.84 & 0.86 & 0.81 & \textbf{\textcolor{black}{0.77}} & \textbf{\textcolor{black}{0.74}} & 0.81 & 0.64 & 0.66 & 0.76 & 0.70 & 0.89 \\
    DPCIAN \cite{yang2023dpcian} & 2023 & 0.91 & 0.88 & 0.83 & 0.83 & 0.79 & \textbf{\textcolor{black}{0.89}} & 0.77 & 0.59 & 0.61 & 0.58 & \underline{0.71} & 0.66 & 0.78 & \textbf{\textcolor{black}{0.77}} & 0.75 \\
    PIT \cite{zhou_pit_2023} & 2023 & 0.91 & 0.90 & 0.82 & \underline{0.85} & 0.79 & 0.87 & \underline{0.87} & 0.66 & 0.54 & \textbf{\textcolor{black}{0.85}} & 0.70 & 0.65 & \underline{0.81} & 0.71 & \underline{0.93} \\
    PedAST-GCN \cite{ling2024pedast} & 2024 & 0.91 & \textbf{\textcolor{black}{0.94}} & 0.83 & \textbf{\textcolor{black}{0.88}} & 0.79 & \textbf{\textcolor{black}{0.89}} & 0.83 & 0.68 & \underline{0.67} & 0.69 & 0.69 & 0.66 & 0.79 & 0.68 & \underline{0.93} \\
    \midrule
    TrajFusionNet-Small (ours) & 2025 & 0.91 & 0.90 & 0.85 & 0.82 & \underline{0.89} & 0.86 & 0.81 & 0.65 & 0.59 & 0.73 & 0.69 & 0.62 & 0.79 & 0.69 & \underline{0.93} \\
    TrajFusionNet (ours) & 2025 & \textbf{\textcolor{black}{0.92}} & \underline{0.91} & \textbf{\textcolor{black}{0.86}} & \underline{0.85} & 0.88 & \textbf{\textcolor{black}{0.89}} & 0.86 & \underline{0.72} & 0.64 & 0.82 & \textbf{\textcolor{black}{0.74}} & \underline{0.68} & \textbf{\textcolor{black}{0.82}} & 0.73 & \underline{0.93} \\ 
    \bottomrule  
  \end{tabular}
  }
  \label{tab:comparison_sota}
\end{table*}

\subsection{Inference Time}

Table~\ref{tab:inference_time} compares the inference times of TrajFusionNet with several SOTA approaches. The inference times were measured on a consumer-grade GPU (NVIDIA GeForce RTX 3060). Only approaches for which the authors have provided the source code are included. Two inference times are reported: the inference time of the model alone (M) and the inference time of the model combined with data preprocessing (M + D). 

Data preprocessing includes the time required to compute modalities required by the models, such as pose estimation and segmentation maps. For instance, PCPA \cite{kotseruba_4_benchmark_2021} uses OpenPose \cite{cao2017realtime} to compute pose estimation, which takes 32.98 ms to run on an RTX 3060 GPU. Yang et al. \cite{yang_2_predicting_2022} and Pedestrian Graph + \citep{cadena2022pedestrian} utilize both OpenPose for pose estimation and DeepLabV3 \cite{chen2017rethinking} for computing a segmentation map. DeepLabV3 requires 157.30 ms to execute on the RTX 3060 GPU.

Table~\ref{tab:inference_time} shows that although TrajFusionNet has a relatively large number of parameters (58.26M), its total inference time (M + D) is 12.09 ms, which is by far the lowest when compared to the other approaches. TrajFusionNet only requires three modalities (pedestrian bounding boxes, vehicle speed, and scene images), meaning it does not spend time computing computationally expensive modalities such as pose estimation and segmentation maps. The lightweight version of TrajFusionNet, TrajFusionNet-Small, has 10 times fewer parameters (5.20M) and reduces the total inference time (M + D) to only 4.63 ms.

\begin{table}
  \caption{Inference time obtained by state-of-the-art models and their respective number of parameters. M represents the inference time of the model alone, while M + D represents the inference time of the model combined with data preprocessing.}
  \resizebox{\columnwidth}{!}{%
  \begin{tabular}{lccc}
    \toprule
    \multirow{2}{*}{Model} & \multicolumn{2}{c}{Inference time (ms)} & \multirow{2}{*}{\makecell{Params \\ (millions)}} \\
    \cmidrule(lr){2-3}
                                & M & M + D \\ 
    \toprule
    PCPA \cite{kotseruba_4_benchmark_2021} & 15.51 & 48.49 & 31.17 \\
    Yang et al. \cite{yang_2_predicting_2022} & 2.64 & 192.92 & 2.99 \\
    Pedestrian Graph + \citep{cadena2022pedestrian} & 2.67 & 192.95 & 0.07 \\
    \midrule
    TrajFusionNet-Small (ours) & 4.63 & 4.63 & 5.20 \\
    TrajFusionNet (ours) & 12.09 & 12.09 & 58.26 \\
    \bottomrule  
  \end{tabular}
  }
  \label{tab:inference_time}
\end{table}

\subsection{Ablation Study}

An ablation study was performed to identify how different components of TrajFusionNet contribute to the model performance. Table~\ref{tab:ablation_study} shows different scenarios in which one component of the network was removed or modified. 

In Scenario 1, instead of merging the outputs from the SAM and VAM branches using a dense layer, the outputs are merged with a modality self-attention layer\footnote[1]{The output from the SAM branch, \begin{math}\boldsymbol{\chi}_{traj}\end{math}, is first concatenated with the output of the VAM branch, \begin{math}\boldsymbol{\chi}_{visual}\end{math}, to form \begin{math}\boldsymbol{\chi}_{combined}\in \mathbb{R}^{d_{\text{mod}} \times d_{\text{enc}}}\end{math}, with \begin{math}d_{mod}=2\end{math} and \begin{math}d_{enc}=40\end{math}. Attention is then computed along the \begin{math}d_{mod}\end{math} dimension using scaled dot product attention \cite{vaswani_attention_2017}. The computed attention context is concatenated with \begin{math}\boldsymbol{\chi}_{combined}\end{math}, and the result is then passed to the final dense layer.}. Late fusion with an attention layer has been employed in previous works \citep{kotseruba_4_benchmark_2021, gesnouin2021trouspi,lorenzo_4_capformer_2021, yang_2_predicting_2022}. However, as shown in Scenario 1 of Table~\ref{tab:ablation_study}, no improvement is observed in TrajFusionNet when merging the two branches with a modality self-attention layer.

In Scenario 2, the number of VAN networks forming the VAM branch is reduced from two to one. The single VAN is applied to the last frame available in the observation period, \begin{math}\boldsymbol{I}^{t}\end{math}. The pedestrian bounding box overlays for the observed trajectory (\begin{math}\boldsymbol{M}^{t-14:t}_{1:4}\end{math}) and the predicted trajectory (\begin{math}\boldsymbol{\hat{M}}^{t+1:t+61}_{1:4}\end{math}) are combined into the same image. Using a single VAN network instead of two does not affect the results on the PIE dataset but leads to a decrease in several metrics on the JAAD{\scriptsize all} and JAAD{\scriptsize beh} datasets

In Scenario 3 of Table~\ref{tab:ablation_study}, sequence type identifiers are removed before sending the past trajectory tensor  (\begin{math}\boldsymbol{M}^{t-14:t}\end{math}) and the predicted trajectory tensor (\begin{math}\boldsymbol{\hat{M}}^{t+1:t+61}\end{math}) to the encoder transformer. A slight decrease in performance across metrics and datasets is observed. In Scenario 4, vehicle speed is removed from the input modalities in the SAM branch (the trajectory prediction transformer only predicts future bounding boxes). Removing the vehicle speed leads to a noticeable decrease in performance across metrics on the PIE and JAAD{\scriptsize beh} datasets but does not affect results significantly on JAAD{\scriptsize all}.

We conclude with Scenarios 5 and 6, where trajectory prediction is excluded from the SAM and VAM branches respectively. In Scenario 5, the predicted trajectory tensor (\begin{math}\boldsymbol{\hat{M}}^{t+1:t+61}\end{math}) is removed from the input to the encoder transformer. As shown in Table~\ref{tab:ablation_study}, this leads to a significant decrease in metrics on both the PIE and JAAD{\scriptsize beh} datasets, and a smaller decrease on the JAAD{\scriptsize all} dataset. In Scenario 6, the predicted trajectory tensor (\begin{math}\boldsymbol{\hat{M}}^{t+1:t+61}_{1:4}\end{math}) is not used to add predicted pedestrian boxes as overlays on the scene image. Instead, a single VAN network is used with overlays from the observed trajectory only (\begin{math}\boldsymbol{M}^{t-14:t}_{1:4}\end{math}). While the results on the PIE dataset remain unaffected, there is a significant decrease in performance on the JAAD{\scriptsize all} and JAAD{\scriptsize beh} datasets.

\begin{table*}[t]
  \centering
  \caption{Ablation study where the proposed TrajFusionNet architecture is compared with various architectural modifications.}
  \resizebox{\textwidth}{!}{%
  \begin{tabular}{clccccccccccccccc}
    \toprule
    \multicolumn{2}{c}{\multirow{2}{*}{Scenario}} & \multicolumn{5}{c}{PIE} & \multicolumn{5}{c}{JAAD{\scriptsize all}} & \multicolumn{5}{c}{JAAD{\scriptsize beh}} \\
    \cmidrule(lr){3-7} \cmidrule(lr){8-12} \cmidrule(lr){13-17}
    & & Acc & AUC & F1 & P & R & Acc & AUC & F1 & P & R & Acc & AUC & F1 & P & R \\ 
    \toprule
    base & \makecell[l]{TrajFusionNet architecture}  & 0.92 & 0.91 & 0.86 & 0.85 & 0.88 & 0.89 & 0.86 & 0.72 & 0.64 & 0.82 & 0.74 & 0.68 & 0.82 & 0.73 & 0.93 \\ \hline
    1 & \makecell[l]{Combine branches with a \\ modality self-attention layer} & 0.92 & 0.91 & 0.86 & 0.82 & 0.90 & 0.88 & 0.84 & 0.70 & 0.64 & 0.77 & 0.71 & 0.64 & 0.81 & 0.70 & 0.95 \\ \hline
    2 & \makecell[l]{Use a single VAN network \\ with combined overlays of \\ past and predicted trajectories} & 0.92 & 0.91 & 0.86 & 0.83 & 0.89 & 0.88 & 0.81 & 0.67 & 0.64 & 0.70 & 0.73 & 0.67 & 0.81 & 0.73 & 0.91 \\ \hline
    3 & \makecell[l]{Remove type IDs in SAM \\branch} & 0.92 & 0.90 & 0.86 & 0.85 & 0.86 & 0.88 & 0.82 & 0.69 & 0.65 & 0.72 & 0.73 & 0.66 & 0.81 & 0.71 & 0.94 \\ \hline
    4 & \makecell[l]{Remove vehicle speed \\ from input modalities} & 0.88 & 0.86 & 0.79 & 0.78 & 0.80 & 0.89 & 0.83 & 0.70 & 0.66 & 0.75 & 0.70 & 0.62 & 0.80 & 0.69 & 0.95 \\ \hline
    5 & \makecell[l]{Remove trajectory prediction \\ from SAM branch} & 0.89 & 0.88 & 0.82 & 0.80 & 0.83 & 0.89 & 0.81 & 0.69 & 0.68 & 0.70 & 0.70 & 0.63 & 0.79 & 0.70 & 0.91 \\ \hline
    6 & \makecell[l]{Remove trajectory prediction \\ from VAM branch} & 0.92 & 0.91 & 0.86 & 0.83 & 0.89 & 0.87 & 0.79 & 0.64 & 0.61 & 0.67 & 0.70 & 0.63 & 0.80 & 0.70 & 0.93 \\
    \bottomrule  
  \end{tabular}
  }
  \label{tab:ablation_study}
\end{table*}

\subsection{Qualitative Results}

\begin{figure*}
    \begin{center}
        \includegraphics[width=0.8\linewidth]{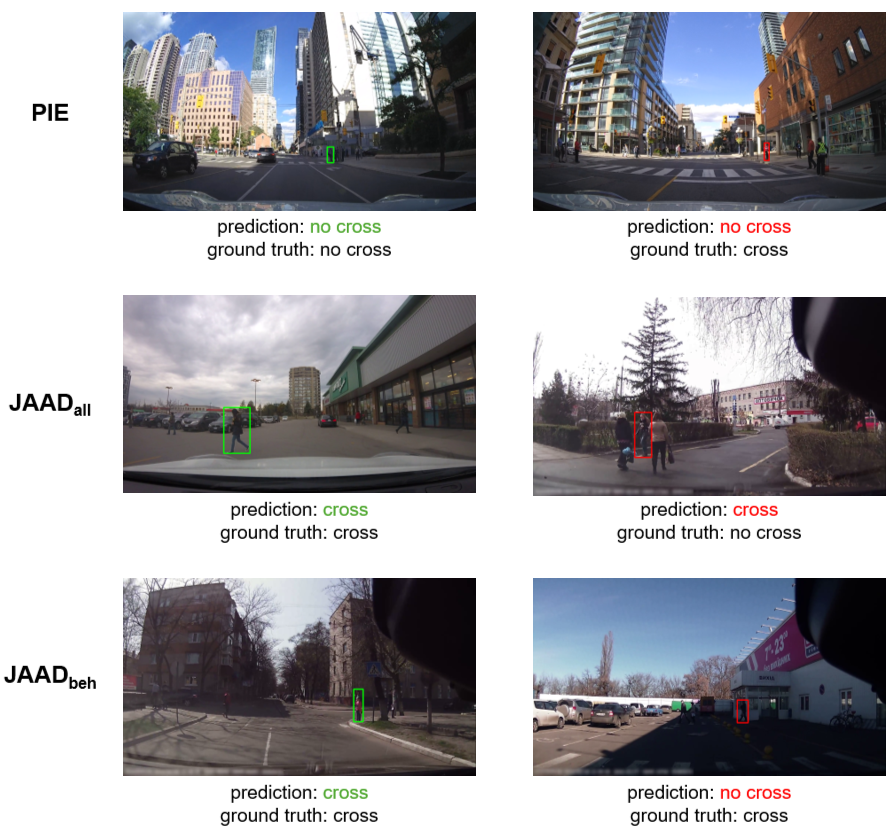}
    \end{center}
   \vspace*{-10pt}
   \caption{Qualitative results showing examples of correct predictions by TrajFusionNet (left column) and incorrect predictions (right column)}
\label{fig:figure_qualitative_analysis}
\end{figure*}

In this section, we analyze examples of correct and incorrect predictions by TrajFusionNet across the three datasets (Figure~\ref{fig:figure_qualitative_analysis}). The left column of the figure presents cases where TrajFusionNet made correct predictions, while the right column shows examples of incorrect predictions. For the PIE dataset, the incorrect example (first row, second column) involves a pedestrian positioned near but not directly in front of a zebra crossing. This seems to mislead the model into inferring that the pedestrian does not intend to cross yet. Additionally, the pedestrian's trajectory during the observation period is relatively static. The pedestrian quickly decides to jaywalk (bypassing the zebra crossing) after the vehicle stops at the crossing to allow the other pedestrians to cross. The model incorrectly predicts "no cross". For the JAAD{\scriptsize all} dataset, the incorrect example (second row, second column) depicts a vehicle turning at an intersection. Due to the turn, the target pedestrian's trajectory appears to move horizontally over time, resembling the trajectory of a crossing pedestrian. However, the pedestrian is merely walking along the roadside. TrajFusionNet incorrectly predicts "cross". For the JAAD{\scriptsize beh} dataset, the incorrect example (third row, second column) shows a pedestrian exiting a store and turning perpendicularly to cross the road, which is an untypical trajectory. Low illumination around the pedestrian further complicates the VAM branch's ability to perceive the scene. TrajFusionNet predicts "no cross", but the pedestrian ultimately crosses the road.

\section*{Conclusion}

This work introduced TrajFusionNet, a novel model for predicting pedestrian crossing intention. The model combines future pedestrian trajectory and vehicle speed predictions as priors for predicting crossing intention. The Sequence Attention Module (SAM) processes a sequential representation of past and future trajectories, while the Visual Attention Module (VAM) utilizes a visual representation of the pedestrian trajectories by overlaying observed and predicted bounding boxes onto scene images. TrajFusionNet achieves state-of-the-art performance across the three most widely used datasets for pedestrian crossing intention prediction. Moreover, by employing a small number of lightweight modalities, TrajFusionNet achieves the lowest total inference time (including model runtime and data preprocessing) among current state-of-the-art approaches.

\section*{Acknowledgments}

This research was enabled in part by support provided by the Natural Sciences and Engineering Research Council of Canada (NSERC), funding reference number RGPIN-2024-05287.

{\small
\bibliographystyle{IEEEtran}
\bibliography{main}
}

\end{document}